\documentclass[runningheads]{llncs}

\usepackage{graphicx}
\usepackage{amsmath}
\usepackage{pdfpages}
\usepackage{color}

\usepackage{thmtools, thm-restate}

\usepackage{booktabs}

\usepackage{epsfig}
\usepackage{dsfont}
\usepackage{cite}
\usepackage{amsmath}
\usepackage{amssymb}
\usepackage{multirow}

\usepackage[normalem]{ulem}
\usepackage{url}
\usepackage{dirtytalk}

\usepackage[normalem]{ulem}
\useunder{\uline}{\ul}{}

\usepackage[subrefformat=parens,labelformat=parens]{subcaption}
\usepackage[pagebackref,breaklinks,colorlinks]{hyperref}
\usepackage[misc]{ifsym}
\usepackage{bbding}
\usepackage{float}

\begin{document}
\title{NVUM: Non-Volatile Unbiased Memory for Robust Medical Image Classification}
\author{
Fengbei Liu \inst{1}\Envelope $\quad$
Yuanhong Chen \inst{1} $\quad$
Yu Tian\inst{1}$\quad$
Yuyuan Liu \inst{1} $\quad$ \\
Chong Wang \inst{1} $\quad$
Vasileios Belagiannis \inst{2}$\quad$ 
Gustavo Carneiro\inst{1}
}%

\institute{Australian Institute for Machine Learning, University of Adelaide \and
Otto von Guericke University Magdeburg 
}
\maketitle              %

\begin{abstract}

Real-world large-scale medical image analysis (MIA) datasets have three challenges: 
1) they contain noisy-labelled samples that affect training convergence and generalisation, 
2) they usually have an imbalanced distribution of samples per class, and 
3) they normally comprise a multi-label problem, where samples can have multiple diagnoses. 
Current approaches are commonly trained to solve a subset of those problems, but we are unaware of methods that address the three problems simultaneously.
In this paper, we propose a new training module called Non-Volatile Unbiased Memory (NVUM), which non-volatility stores running average of model logits for a new regularization loss on noisy multi-label problem. We further unbias the classification prediction in NVUM update for imbalanced learning problem.
We run extensive experiments to evaluate NVUM on new benchmarks proposed by this paper, where training is performed on noisy multi-label imbalanced chest X-ray (CXR) training sets, formed by Chest-Xray14 and CheXpert, and the testing is performed on the clean multi-label CXR datasets OpenI and PadChest. 
Our method outperforms previous state-of-the-art CXR classifiers and previous methods that can deal with noisy labels on all evaluations. Our code is available at \url{https://github.com/FBLADL/NVUM}. \footnote{This work was supported by the Australian Research Council through grants DP180103232 and FT190100525.}

\end{abstract}

\section{Introduction and Background}

The outstanding results shown by deep learning models in medical image analysis (MIA)~\cite{litjens2017survey,liu2021acpl}
depend on the availability of large-scale manually-labelled training sets, which is expensive to obtain. 
As a affordable alternative, these  manually-labelled training sets can be replaced by datasets that are automatically labelled by natural language processing (NLP) tools that extract labels from the radiologists' reports~\cite{wang2017chestx,irvin2019chexpert}. 
However, the use of these alternative labelling processes often produces unreliably labelled datasets because NLP-extracted disease labels, without verification by doctors, may contain incorrect labels, which are called \textit{noisy labels}~\cite{oakden2017exploring,oakden2020exploring}. 
Furthermore, differently from computer vision problems that tend to be multi-class with a balanced distribution of samples per class, MIA problems are usually multi-label (e.g, a disease sample can contain multiple diagnosis), with severe class imbalances because of the variable prevalence of diseases. 
Hence, robust MIA methods need to be flexible enough to work with \textit{noisy multi-label} and \textit{imbalanced} problems.

\begin{figure}[t]
    \centering
    \includegraphics[width=0.95\textwidth]{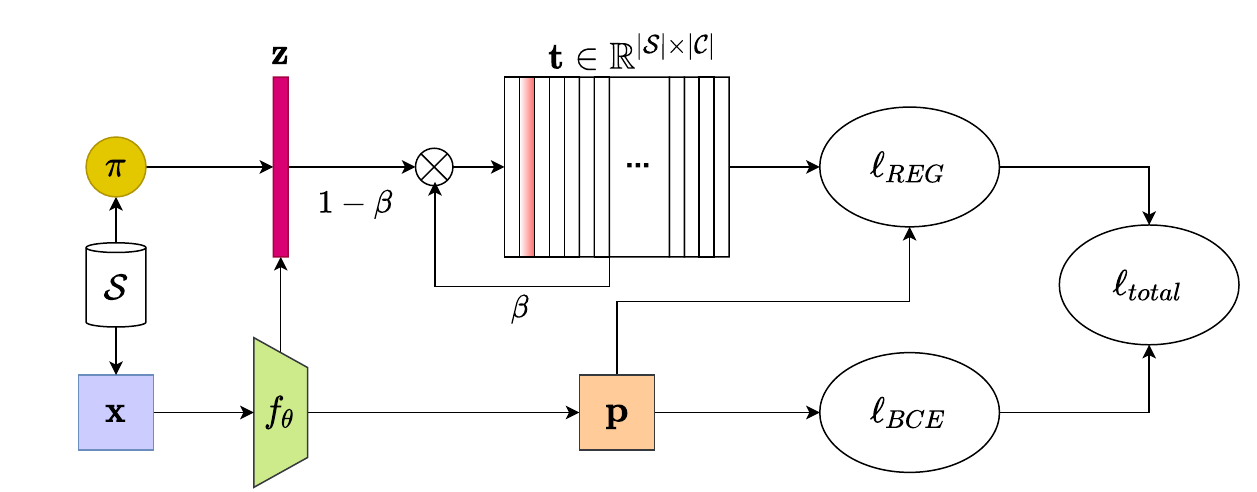}
    \caption{NVUM training algorithm: 1) sample input image $\mathbf{x}$ from training set $\mathcal{S}$ and calculate label distribution prior $\pi$; 2) train model $f_\theta$ and get sample logits $\mathbf{z}$ and prediction $\mathbf{p}$; 3) update memory $\mathbf{t}$ with~\eqref{eq:memory_module}; and 4) minimise the loss that comprises $\ell_{BCE}(.)$ in~\eqref{eq:main_loss} and $\ell_{REG}(.)$ in~\eqref{eq:ell_REG}.}  
    \label{fig:nvum}
\end{figure}
State-of-the-art (SOTA) noisy-label learning approaches are usually based on noise-cleaning methods~\cite{li2020dividemix,liu2020early,han2018co}. Han et al.~\cite{han2018co} propose to use two DNNs and use their disagreements to reject noisy samples from the training process. Li et al.~\cite{li2020dividemix} rely on semi-supervised learning that treats samples classified as noisy as unlabelled samples. 
Other approaches estimate the label transition matrix~\cite{goldberger2016training,xia2019anchor} to correct model prediction. %
Even though these methods show state-of-the-art (SOTA) results in noisy-label problems, they have issues with imbalanced and multi-label problems. 
First, noise-cleaning methods usually rely on detecting noisy samples by selecting large training loss samples, which are either removed or re-labelled.
However, in imbalanced learning problems, such training loss for clean-label training samples, belonging to minority classes, can be larger than the loss for noisy-label training samples belonging to majority classes, so
these SOTA noisy-label learning approaches may inadvertently remove or re-label samples belonging to minority classes. 
Furthermore, in multi-label problems, the same sample can have a mix of clean and noisy labels, so it is hard to adapt SOTA noisy-label learning approaches to remove or re-label particular labels of each sample.
Another issue in multi-label problems faced by transition matrix methods is that they are designed to work for multi-class problems, so their adaptation to multi-label problems will need to account for the correlation between the multiple labels.
Hence, current noisy-label learning approaches have not been designed to solve all issues present in noisy multi-label imbalanced real-world datasets.

Current imbalanced learning approaches are usually based on decoupling  classifier and representation learning~\cite{kang2019decoupling,tang2020longtailed}. 
For instance, Kang et al.~\cite{kang2019decoupling} 
notice that learning with an imbalanced training set does not affect the representation learning, so they only adjust for imbalanced learning when training the classifier.
Tang et al.~\cite{tang2020longtailed} identify causal effect in stochastic gradient descent (SGD) momentum update on imbalanced datasets and propose a de-confounded training scheme. 
Another type of imbalanced learning is based on loss weighting~\cite{cao2019learning,tan2020equalization} that up-weights the minority classes~\cite{cao2019learning} or down-weights the majority classes~\cite{tan2020equalization}.  
Furthermore, Menon et al.~\cite{menon2020long} discover that decoupling approach that based on correlation between classifier weight norm and data distribution is only applicable for SGD optimizer,
which is problematic for MIA methods that tend to rely on other optimizers, such as Adam, that show better training convergence. 
Even though the papers above are effective for imbalanced learning problems, they do not consider the combination of imbalanced and noisy multi-label learning.

To address the noisy multi-label imbalanced learning problems present in real-world MIA datasets, we introduce the \textbf{Non-volatile Unbiased Memory (NVUM)} training module, which is described in Fig.~\ref{fig:nvum}. Our contributions are: 
\begin{itemize}
    \item NVUM that stores a non-volatile running average of model logits to explore the multi-label noise robustness of the early learning stages. This memory module is used by a new regularisation loss to penalise differences between current and early-learning model logits;
    \item The NVUM update takes into account the class prior distribution to unbias the classification predictions estimated from the imbalanced training set;  
    \item Two new noisy multi-label imbalanced evaluation benchmarks, where training is performed on chest X-ray (CXR) training sets from Chest Xray14~\cite{wang2017chestx} and CheXpert~\cite{irvin2019chexpert}, and testing is done on the clean multi-label CXR datasets OpenI~\cite{demner2016preparing} and PadChest~\cite{bustos2020padchest}.
\end{itemize}

\section{Method}
\label{sec:method}

We assume the availability of a noisy-labelled training set $\mathcal{S}=\{(\mathbf{x}^i,\mathbf{\tilde{y}}^i)\}_{i=1}^{|\mathcal{S}|}$, where $\mathbf{x}^i \in \mathcal{X} \subset \mathbb{R}^{H \times W \times R}$ is the input image of size $H \times W$ with $R$ colour channels, and $\mathbf{\tilde{y}}^i  \in \{0,1\}^{|\mathcal{C}|}$ is the noisy label with the set of classes denoted by $\mathcal{C}=\{1,...,|\mathcal{C}|\}$ (note that $\mathbf{\tilde{y}}_i$ represents a binary vector in multi-label problems, with each label representing one disease).

\subsection{Non-volatile Unbiased Memory (NVUM) Training}
\label{sec:NVUM}
To describe the NVUM training, we first need to define the model, parameterised by $\theta$ and represetned as a deep neural network, with $\mathbf{p} = \sigma(f_{\theta}(\mathbf{x}))$
, where $\mathbf{p} \in [0,1]^{|\mathcal{C}|}$, $\sigma(.)$ denotes the sigmoid activation function and
$\mathbf{z} = f_{\theta}(\mathbf{x})$, with $\mathbf{z} \in \mathcal{Z} \in \mathbb{R}^{|\mathcal{C}|}$  representing a logit.
The training of the model $f_{\theta}(\mathbf{x})$ is achieved by minimising the following loss function: 
\begin{equation}
    \ell_{total}(\mathcal{S},\mathbf{t},\theta)=
    \frac{1}{|\mathcal{S}|} \sum_{(\mathbf{x}^{i},\mathbf{\tilde{y}}^{i}) \in \mathcal{S}} \ell_{BCE}(\mathbf{\tilde{y}}^{i},\mathbf{p}^{i}) +  \ell_{REG}(\mathbf{t}^i,\mathbf{p}^i),
    \label{eq:main_loss}
\end{equation}
where $\ell_{BCE}$ denotes the binary cross-entropy loss for handling multi-label classification and $\ell_{REG}$ is a regularization term defined by:
\begin{equation}
\begin{split}
    \ell_{REG}(\mathbf{t}^i,\mathbf{p}^i) &=
    \log(1 - \sigma(
    (\mathbf{t}^i)^{\top}\mathbf{p}^i)).
\end{split}
\label{eq:ell_REG}
\end{equation}
here $\mathbf{t} \in \mathbb{R}^{|\mathcal{S}| \times |\mathcal{C}|}$ is our proposed memory module designed to store an unbiased multi-label running average of the predicted logits for all training samples and $\mathbf{t}$ 
uses the class prior distribution $\pi$ for updating, denoted by $\pi(c) = \frac{1}{|\mathcal{S}|}\sum_{i=1}^{|\mathcal{S}|} \tilde{\mathbf{y}}(c)$ for $c \in \{1,...,|\mathcal{C}|\}$. The memory module $\mathbf{t}$ is initialised with zeros, as in $\mathbf{t}_{0} = \mathbf{0}^{|\mathcal{S}| \times |\mathcal{C}|}$, and is updated in every epoch $k > 0$ with:
\begin{equation}
    \mathbf{t}^{i}_{k} = \beta \mathbf{t}^{i}_{k-1} + (1-\beta)(\mathbf{z}_{k}^{i} - \log \pi),
    \label{eq:memory_module}
\end{equation}
where $\beta \in [0,1]$ is a hyper-parameter controlling the volatility of the memory storage, with $\beta$ set to larger value representing a non-volatile memory and $\beta \approx 0$ denoting a volatile memory that is used in ~\cite{he2019moco} for contrastive learning.
To explore the early learning phenomenon, we set $\beta = 0.9$ so the regularization can enforce the consistency between the current model logits and the logits produced at the beginning of the training, when the model is robust to noisy label.
Furthermore, to make the training robust to imbalanced problems, we subtract the log prior of the class distributions, which has the effect of increasing the logits with larger values for the classes with smaller prior.
This counterbalances the issue faced by imbalanced learning problems, where 
the logits for the majority classes can overwhelm those from the minority classes, to the point that logit inconsistencies found by the regularization from noisy labels of the majority classes may become indistinguishable from the clean labels from minority classes. \\
The effect of Eq.~\eqref{eq:ell_REG} can be interpreted by inspecting the loss gradient, which is proved in the supplementary material. The gradient of~\eqref{eq:main_loss} is:
\begin{equation}
    \begin{split}
        \nabla_{\theta} \ell_{total}(\mathcal{S},\theta) &=\frac{1}{|\mathcal{S}|}
        \sum_{i \in \mathcal{S}} 
        \mathbf{J}_{\mathbf{x}^i}(\theta)(\mathbf{p}^i - \mathbf{\tilde{y}}^i + \mathbf{g}^i) ,\\
        \text{where} \quad 
        \mathbf{g}_c^i &= -\sigma((\mathbf{t}^i)^{\top}(\mathbf{p}^i))\mathbf{p}_c^i(1-\mathbf{p}_c^i)\mathbf{t}_c 
    \end{split}
    \label{eq:gradient}
\end{equation}
where 
$\mathbf{J}_{\mathbf{x}^i}(\theta)$ is the Jacobian matrix w.r.t. $\theta$ for the $i^{th}$ sample. 
Assume $\mathbf{y}_c$ is the hidden true label of the sample $\mathbf{x}^i$, then the entry $\mathbf{t}^i_c > 0$ if $\mathbf{y}_c=1$, and $\mathbf{t}^i_c < 0$ if $\mathbf{y}_c = 0$ at the early stages of training. During training, we consider four conditions explained below, where we assume that $\sigma((\mathbf{t}^i)^{\top}(\mathbf{p}^i))\mathbf{p}_c^i(1-\mathbf{p}_c^i)>0$. When the training sample has clean label:
\begin{itemize}
    \item if $\tilde{\mathbf{y}}_c=\mathbf{y}_c=1$, the gradient of the BCE term,  $\mathbf{p}_c^i - \mathbf{\tilde{y}}_c^i \approx 0$ given that the model is likely to fit clean samples. With $\mathbf{t}^i_c > 0$, the sign of $\mathbf{g}_c^i$ is negative, and the model keeps training for these positive labels even after the early-training stages.
    \item if $\tilde{\mathbf{y}}_c=\mathbf{y}_c=0$, the gradient of the BCE term,  $\mathbf{p}_c^i - \mathbf{\tilde{y}}_c^i \approx 0$ given that the model is likely to fit clean samples. Given that $\mathbf{t}^i_c < 0$, we have $\mathbf{g}_c^i > 0$, and the model keeps training for these negative labels even after the early-training stages.
\end{itemize}
Therefore, adding $\mathbf{g}_c^i$ in total loss ensures that clean samples gradient magnitudes remains relatively high, encouraging a continuing optimisation using the clean label samples. For a noisy-label sample, we have:
\begin{itemize}
    \item if $\tilde{\mathbf{y}}_c = 0$ and $\mathbf{y}_c=1$, the gradient of the BCE loss is $\mathbf{p}_c^i - \mathbf{\tilde{y}}_c^i \approx 1$ because the model will not fit noisy label during early training stages. With $\mathbf{t}^i_c > 0$, we have  $\mathbf{g}_c^i < 0$, which reduces the gradient magnitude from the BCE loss.
    \item if $\tilde{\mathbf{y}}_c = 1$ and $\mathbf{y}_c=0$, $\mathbf{p}_c^i-\mathbf{\tilde{y}}_c^i \approx -1$. Given that $\mathbf{t}^i_c < 0$, we have $\mathbf{g}_c^i > 0$, which also reduces the gradient magnitude from the BCE loss.
\end{itemize}
Therefore, for noisy-label samples, $\mathbf{g}_c^i$ will counter balance the gradient from the BCE loss and diminish the effect of noisy-labelled samples in the training.

\section{Experiment}

\noindent \textbf{Datasets.} For the experiments below, we use the  NIH Chest X-ray14~\cite{wang2017chestx} and CheXpert~\cite{irvin2019chexpert} as noisy multi-label imbalanced datasets for training and  Indiana OpenI~\cite{demner2016preparing} and PadChest~\cite{bustos2020padchest} datasets for clean multi-label testing sets. 

For the noisy sets, \textbf{NIH Chest X-ray14 (NIH)} contains 112,120 CXR images from 30,805 patients. There are 14 labels (each label is a disease), where each patient can have multiple diseases, forming a multi-label classification problem. 
For a fair comparison with previous papers~\cite{hermoza2020region,rajpurkar2017chexnet}, we adopt the official train/test data split~\cite{wang2017chestx}.
\textbf{CheXpert (CXP)} contains 220k images with 14 different diseases, and similarly to NIH, each patient can have multiple diseases. For pre-processing, we remove all lateral view images and treat uncertain and empty labels as negative labels. Given that the clean test set from CXP is not available and the clean validation set is too small for a fair comparison, we further split the training images into 90\% training set and 10\% \textit{noisy} validation set with no patient overlapping.
For the clean sets, \textbf{Indiana OpenI (OPI)} contains 7,470 frontal/lateral images with manual annotations. In our experiments, we only use 3,643 frontal view of images for evaluation. 
\textbf{PadChest (PDC)} is a large-scale dataset containing 158,626 images with 37.5\% of images manually labelled. In our experiment, we only use the manually labelled samples as the clean test set. 
To keep the number of classes consistent between different datasets, we trim the training and testing sets  based on the shared classes between these datasets \footnote{We include a detailed description based on~\cite{Cohen2020xrv} in the supplementary material.}.

\noindent \textbf{Implementation Details.} We use the ImageNet~\cite{russakovsky2015imagenet} pre-trained DenseNet121~\cite{huang2017densely} as the backbone model for $f_\theta(.)$ on NIH and CXP. 
We use Adam~\cite{kingma2014adam} optimizer with batch size 16 for NIH and 64 for CXP. 
For NIH, we train for 30 epochs with a learning rate of 0.05 and decay with 0.1 at 70\% and 90\% of the total of  training epochs. 
Images are resized from 1024$\times$1024 to 512$\times$512 pixels.
For data augmentation, we employ random resized crop and random horizontal flipping. 
For CXP, we train for 40 epochs with a  learning rate of $1e^{-4}$ and follow the learning rate decay policy as on NIH. 
Images are resized to 224$\times$224.
For data augmentation, we employ random resized crop, 10 degree random rotation and random horizontal flipping. 
For both datasets, we use $\beta = 0.9$ and normalized by ImageNet mean and standard deviation.

All classification results are reported  using area under the ROC curve (AUC). 
To report performance on clean test sets OPI and PDC, we adopt a common noisy label setup~\cite{li2020dividemix,han2018co} that selects the best performance checkpoint on noisy validation, which is the noisy test set of NIH and the noisy validation set of CXP. All experiments are implemented with Pytorch~\cite{paszke2019pytorch} and conducted on an NVIDIA RTX 2080ti GPU. The training takes 15 hours on NIH and 14 hours on CXP.

\begin{table}[t]
\centering
\caption{Class-level and mean testing AUC on OPI~\cite{demner2016preparing} and PDC~\cite{bustos2020padchest} for the experiment based on training on NIH~\cite{wang2017chestx}. Best results for OPI/PDC are in \textbf{bold}/\underline{underlined}.}
\label{tab:nih_result}
    \resizebox{0.9\textwidth}{!}{  
    \begin{tabular}{l|cc|cc|cc|cc|cc}
    \toprule
    Models	&	\multicolumn{2}{c|}{ChestXNet~\cite{rajpurkar2017chexnet}}			&	\multicolumn{2}{c|}{Hermoza et al.~\cite{hermoza2020region}}			&	\multicolumn{2}{c|}{Ma et al.~\cite{ma2019multi}}			&	\multicolumn{2}{c|}{DivideMix~\cite{li2020dividemix}}			&	\multicolumn{2}{c}{Ours}			\\	\hline
Datasets            & OPI          & PDC          & OPI             & PDC            & OPI             & PDC             & OPI           & PDC           & OPI         & PDC         \\ \hline
Atelectasis         & 86.97        & 84.99        &    86.85             &   83.59             & 84.83           & 79.88           &     70.98          &    73.48           & \textbf{88.16}       & \underline{85.66}       \\
Cardiomegaly        & 89.89        & 92.50         &   89.49              &   91.25             & 90.87           & 91.72           &     74.74          &     81.63          & \textbf{92.57}       & \underline{92.94}       \\
Effusion            & 94.38        & 96.38        &    95.05             &   96.27             & 94.37           & 96.29           &     84.49          &      \underline{97.75}         & \textbf{95.64}       & 96.56       \\
Infiltration        & 76.72        & 70.18        &   \textbf{77.48}     &   64.61             & 71.88           & 73.78           &     84.03          &      \underline{81.61}         & 72.48       & 72.51       \\
Mass                & 53.65        & 75.21        &   95.72              &   \underline{86.93}             & 87.47           & 85.81           &     71.31          &      77.41         & \textbf{97.06}       & 85.93       \\
Nodule              & 86.34        & 75.39        &   82.68              &   \underline{75.99}             & 69.71           & 68.14           &      57.45         &       63.89        & \textbf{88.79}       & 75.56       \\
Pneumonia           & \textbf{91.44}        & 76.20         &   88.15              &   75.73             & 84.79           & 76.49           &    64.65           &      72.32         & 90.90        & \underline{82.22}       \\
Pneumothorax        & 80.48        & 79.63        &   75.34              &   74.55              & 82.21           & \underline{79.73}           &    71.56           &     75.46          & \textbf{85.78}       & 79.50        \\
Edema               & 83.73        & \underline{98.07}        &   85.31              &   97.78             & 82.75           & 96.41           &      80.71        &      91.81         & \textbf{86.56}       & 95.70        \\
Emphysema           & 82.37        & 79.10         &   83.26              &   \underline{79.81}             & 79.38           & 75.11           &     54.81          &     59.91          & \textbf{83.70}        & 79.38       \\
Fibrosis            & 90.53        & 96.13        &    86.26            &    96.46            & 83.17           & 93.20            &     76.98          &     84.71          & \textbf{91.67}       & \underline{98.40}        \\
Pleural Thickening & 81.58        & 72.29        &    77.99             &    69.95            & 77.59           & 67.87           &      63.98         &      58.25         & \textbf{84.82}       & \underline{74.80}        \\
Hernia              & 89.82        & 86.72        &   93.90              &   89.29             & 87.37           & 86.87           &      66.34         &     72.11          & \textbf{94.28}       & \underline{93.02}       \\ \hline 
Mean AUC            & 83.69        & 83.29        &   86.01              &    83.25            & 82.80            & 82.41           &     70.92          &     76.18          & \textbf{88.65}       & \underline{85.55}       \\ \hline \bottomrule

\end{tabular}}
\end{table}

\subsection{Experiments and Results}

\textbf{Baselines.} We compared NVUM with several methods, including the CheXNet baseline~\cite{rajpurkar2017chexnet}, Ma et al.'s approach~\cite{ma2019multi} based on a cross-attention network,
the current SOTA for NIH on the official test set is the model by Hermoza et al.~\cite{hermoza2020region} that is a weakly supervised disease classifier that combines region proposal and saliency detection. We also show results from DivideMix~\cite{li2020dividemix}, which uses a noisy-label learning algorithm based on small loss sample selection and semi-supervised learning. DivideMix has the SOTA results in many noisy-label learning benchmarks. All methods are implemented using the same DenseNet121~\cite{huang2017densely} backbone.

\noindent\textbf{Quantitative Comparison}  Table~\ref{tab:nih_result} shows the class-level AUC result for training on NIH and testing on OPI and PDC.
Our approach achieves the SOTA results on both clean test sets, consistently outperforming the baselines~\cite{rajpurkar2017chexnet,hermoza2020region,ma2019multi}, achieving 2\% mean AUC improvement on both test sets. 
Compared with the current SOTA noisy-label learning   DivideMix~\cite{li2020dividemix}, our method outperforms it by 18\% on OPI and 9\% on PDC. This shows that for noisy multi-label imbalanced MIA datasets, noisy multi-class balanced approaches based on small-loss selection is insufficient because they do not take into account the multi-label and imbalanced characteristics of the  datasets. 
Table~\ref{tab:cxp_result} shows class-level AUC results for training on CXP and testing on OPI and PDC. Similarly to the NIH results on Table~\ref{tab:nih_result}, our approach achieves the best AUC results on both test sets with at least 3\% improvement on OPI and 3\% on PDC. In addition, DivideMix~\cite{li2020dividemix} shows similar results compared with NIH. Hence, SOTA performance on both noisy training sets suggests that our method is robust to different noisy multi-label imbalanced training sets.

\begin{table}[t]
\centering
\caption{Class-level and mean testing AUC on OPI~\cite{demner2016preparing} and PDC~\cite{bustos2020padchest} for the experiment based on training on CXP~\cite{irvin2019chexpert}.Best results for OPI/PDC are in \textbf{bold}/\underline{underlined}.}
\label{tab:cxp_result}
\resizebox{0.9\textwidth}{!}{
\begin{tabular}{l|cc|cc|cc|cc|cc}
\toprule \hline
Methods             & \multicolumn{2}{c|}{CheXNet ~\cite{rajpurkar2017chexnet}} & \multicolumn{2}{c|}{Hermoza et al.~\cite{hermoza2020region} } & \multicolumn{2}{c|}{Ma et al.~\cite{ma2019multi}} & \multicolumn{2}{c|}{DivideMix\cite{li2020dividemix}} & \multicolumn{2}{c}{Ours} \\ \hline
Datasets     & OPI          & PDC          & OPI             & PDC            & OPI             & PDC             & OPI           & PDC           & OPI         & PDC         \\ \hline
Cardiomegaly & 84.00           & 80.00           &  87.01               &  87.20              &     82.83            &    85.89             &    71.14           &   66.51            & \textbf{88.86}       & \underline{88.48}       \\
Edema        & 88.16        & 98.80         &  87.92               &   98.72             &86.46                      &      97.47           &     75.36         &    95.51           & \textbf{88.63}       & \underline{99.60}        \\
Pneumonia    & \textbf{65.82}        & 58.96        &   65.56              &    53.42            &   61.88              &     54.83            &    57.65           &     40.53          & 64.90        & \underline{67.89}       \\
Atelectasis  & 77.70         & 72.23        &   78.40              &   \underline{75.33}             &  80.13               &    72.87             &    73.65           &    64.12           & \textbf{80.81}       & 75.03       \\
Pneumothorax & 77.35        & \underline{84.75}        &   62.09              &    78.65            &  51.08               &     71.57             &     68.75          &      54.05         & \textbf{82.18}       & 83.32       \\
Effusion     & 85.81        & 91.84        &   87.00              &    \underline{93.44}            &  \textbf{88.43}               &     92.92            &    78.60           &    79.89           & 83.54       & 89.74       \\
Fracture     & 57.64        & 60.26        &   57.47              &    53.77            &   59.92              &    60.44             &    \textbf{60.35}           &    59.43           & 57.02       & \underline{62.67}       \\ \hline
Mean AUC     & 76.64        & 78.12        &  75.06               &    77.29            &   72.96              &    76.57             &    69.36           &    65.72           & \textbf{77.99}       & \underline{80.96}   \\ \hline \bottomrule
\end{tabular}}
\end{table}
\begin{table}[t]
\centering
\caption{Pneumothorax and Mass/Nodule AUC using the manually labelled clean test from~\cite{majkowska2020chest}. Baseline results obtained from~\cite{xue2022robust}. Best results are in \textbf{bold}.}
\label{tab:clean_test_set}
\resizebox{0.9\textwidth}{!}{%
\begin{tabular}{l|ccccccc|c}
\toprule \hline
        & BCE  & F-correction~\cite{patrini2017making} & MentorNet~\cite{jiang2018mentornet} & Decoupling~\cite{malach2017decoupling} & Co-teaching~\cite{han2018co} & ELR~\cite{liu2020early}  & Xue et al.~\cite{xue2022robust} & Ours \\ \hline
Pneu   & 87.0 & 80.8         & 86.6      & 80.1       & 87.3        & 87.1 & \textbf{89.1}       & \textbf{88.9} \\
M/N & 84.3 & 84.8         & 83.7      & 84.3       & 82.0        & 83.2 & 84.6       & \textbf{85.5} \\ \hline \bottomrule
\end{tabular}%
}
\end{table}
\begin{figure}[t]
\raisebox{0.15\height}{
\begin{minipage}[b]{0.4\linewidth}
\centering
\resizebox{1\textwidth}{!}{
\begin{tabular}[b]{cccc|cc}
        \toprule \hline
            $\ell_{BCE}$       &\multicolumn{1}{c|}{$\pi$}& $\ell_{REG}$       & $\pi$     & OPI & PDC \\ \hline
            \checkmark & \multicolumn{1}{c|}{}          &           &           &  82.91$\pm$0.78   &   82.27$\pm$1.02  \\
            \checkmark & \multicolumn{1}{c|}{\checkmark} &           &           &85.80$\pm$0.04    &  84.35$\pm$0.35   \\ 
            \checkmark & \multicolumn{1}{c|}{}          & \checkmark &           &  85.24$\pm$0.70   &  84.39$\pm$0.21   \\ \hline
            \checkmark & \multicolumn{1}{c|}{\checkmark} & \checkmark &           & 85.36$\pm$0.11    &   83.04$\pm$0.79  \\
            \checkmark & \multicolumn{1}{c|}{\checkmark} & \checkmark & \checkmark &  86.68$\pm$0.16   & 85.02$\pm$0.18    \\ \hline
            \multicolumn{4}{c|}{NVUM}                                          &  \textbf{88.17$\pm$0.48}   & \textbf{85.49$\pm$0.06}   \\ \hline \bottomrule
    \end{tabular}}
\end{minipage}}
\hfill
\raisebox{0\height}{
\begin{minipage}[b]{0.45\linewidth}
\centering
\includegraphics[width=0.8\textwidth]{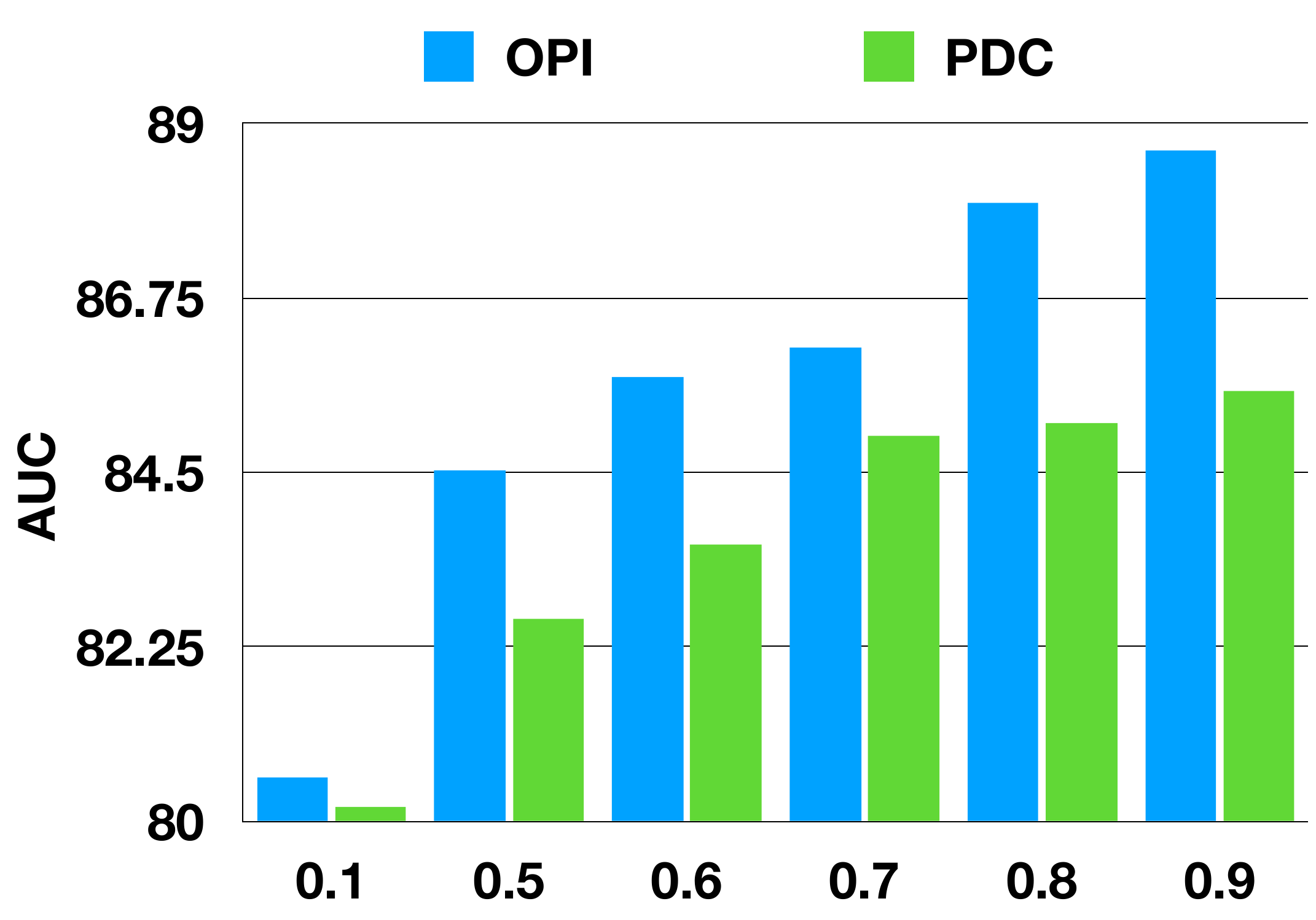}
\end{minipage}}

\caption{(Left) Mean AUC results of training on NIH using the class prior distribution $\pi$ applied to different components of NVUM.  (Right) Mean AUC results on OPI (blue) and PDC (green) of training on NIH with different $\beta$ values for the NVUM memory update in~\eqref{eq:memory_module}.
\label{tab:ablation}}
\end{figure}
\noindent\textbf{Additional benchmark.} Using the recently proposed noisy label benchmark by Xue et al.~\cite{xue2022robust}, we further test our approach against the SOTA in the field.
The benchmark uses a subset of the official NIH test set~\cite{majkowska2020chest}, with 
1,962 CXR images manually re-labelled by at least three radiologists per image.
For the results, we follow~\cite{xue2022robust}
and consider the AUC results only for Pneumothorax (Pneu) and average of Mass and Nodule (M/N). We use the same hyperparameters as above.
The results in Tab.~\ref{tab:clean_test_set} shows that our method outperforms most noisy label methods and achieves comparable performance to~\cite{xue2022robust} on Pneumothorax (88.9 vs 89.1) and better performance on Mass/Nodule (85.5 vs 84.6). 
However, it is important to mention that differently from~\cite{xue2022robust} that uses two models, we use only one model, so our method requires significantly less training time and computation resources. Furthermore, the clean test set from~\cite{majkowska2020chest} is much smaller than OPI and PDC with only two classes available, 
so we consider results in Tab.~\ref{tab:nih_result} and~\ref{tab:cxp_result} more reliable than Tab.~\ref{tab:clean_test_set}

\subsection{Ablation Study}

\noindent \textbf{Different components of NVUM with $\pi$.} We first study in Fig.~\ref{tab:ablation} (left) how results are affected by the prior added on different components of NVUM. 
We run each experiment three times and show mean and standard deviation of AUC results. 
By adding the class prior $\pi$ to $\ell_{BCE}$~\cite{menon2020long}, we replace the BCE term in~\eqref{eq:main_loss} with $\ell_{BCE}(\mathbf{\tilde{y}}^{i},\sigma(f_{\theta}(\mathbf{x}^{i} + \log\pi)))$. We can also add the class prior $\pi$ to
$\ell_{REG}$ by replacing the regularization term in~\eqref{eq:main_loss} with  $\ell_{REG}(\mathbf{t}^i,\sigma(f_{\theta}(\mathbf{x} + \log\pi)))$.
We observe a 2\% improvement for OPI and PDC for both modifications compared to $\ell_{BCE}$ baseline, demonstrating that it is important to handle imbalanced learning in MIA problems. Furthermore, we combine two modifications together and achieve additional 1\% improvement. However, instead of directly working on the loss functions, as suggested in~\cite{menon2020long}, we work on the memory module given that it also enforces the early learning phenomenon, addressing the combined noisy multi-label imbalanced learning problem.\\
\noindent\textbf{Different $\beta$.} We also study different values for $\beta$ in~\eqref{eq:memory_module}. First, we test a volatile memory update with $\beta = 0.1$, which shows a significantly worse performance because  the model is overfitting the noisy multi-label of the training set. This indicates traditional volatile memory~\cite{he2019moco} cannot handle noisy label learning.
Second, the non-volatile memory update with $\beta \in \{0.5,...,0.9\}$ shows a performance that improves consistently with larger $\beta$. Hence, we use $\beta = 0.9$ as our default setup.

\section{Conclusions and Future Work}
In this work, we argue that the MIA problem is a problem of \textit{noisy multi-label} and \textit{imbalanced learning}. We presented the Non-volatile Unbiased Memory (NVUM) training module, which stores a non-volatile running average of model logits to make the learning robust to noisy multi-label datasets. Furthermore, 
The NVUM takes into account the class prior distribution when updating the memory module to make the learning robust to imbalanced learning. We conducted experiments on proposed new benchmark and recent benchmark~\cite{xue2022robust} and achieved SOTA results. Ablation study shows the importance of carefully accounting for imbalanced and noisy multi-label learning. For the future work, we will explore an precise estimation of class prior $\pi$ during the training for accurate unbiasing.

 \newpage

\bibliographystyle{splncs04}
\bibliography{bibi}

\section{Appendix: Gradient Proof}
The gradient for $\ell_{total}(\mathcal{S},\mathbf{t},\theta)=
    \frac{1}{|\mathcal{S}|} \sum_{(\mathbf{x}^{i},\mathbf{\tilde{y}}^{i}) \in \mathcal{S}} \ell_{BCE}(\mathbf{\tilde{y}}^{i},\mathbf{p}^{i}) +  \ell_{REG}(\mathbf{t}^i,\mathbf{p}^i),
    $ is defined as
\begin{equation}
    \begin{split}
        \nabla_{\theta} \ell_{total}(\mathcal{S},\theta) &=\frac{1}{|\mathcal{S}|}
        \sum_{i \in \mathcal{S}} 
        \mathbf{J}_{\mathbf{x}^i}(\theta)(\mathbf{p}^i - \mathbf{\tilde{y}}^i + \mathbf{g}^i) ,\\
        \text{where} \quad 
        \mathbf{g}_c^i &= -\sigma((\mathbf{t}^i)^{\top}(\mathbf{p}^i))\mathbf{p}_c^i(1-\mathbf{p}_c^i)\mathbf{t}_c, 
    \end{split}
\end{equation}
where $\nabla_{\theta}\ell_{BCE}(\theta)  = \frac{1}{|\mathcal{S}|}\sum_{i \in \mathcal{S}} \mathbf{J}_{\mathbf{x}^i}(\theta)(\mathbf{p}^i-\tilde{\mathbf{y}}^i)$, and recalling that $\ell_{REG}(\mathbf{t}^i,\mathbf{p}^i) = \log(1 - \sigma((\mathbf{t}^i)^{\top}\mathbf{p}^i))$ (with $\mathbf{p} =\frac{1}{1+e^{-f_\theta(\mathbf{x})}}$), we have
\begin{equation}
    \begin{split}
        \nabla_{\theta}\ell_{REG}(\theta)^i_c 
        & = \mathbf{g}^i_c \\& =  \frac{1}{1 - \sigma((\mathbf{t}^i)^{\top}(\mathbf{p}^i))}\nabla_{\theta}(1 - \sigma((\mathbf{t}^i)^{\top}(\mathbf{p}^i)))\\
        & = \frac{-1}{1 - \sigma((\mathbf{t}^i)^{\top}(\mathbf{p}^i))}\nabla_{\theta}(\sigma((\mathbf{t}^i)^{\top}(\mathbf{p}^i)))\\
        & = \frac{-1}{1 - \sigma((\mathbf{t}^i)^{\top}(\mathbf{p}^i))}(\sigma((\mathbf{t}^i)^{\top}(\mathbf{p}^i)))(1-\sigma((\mathbf{t}^i)^{\top}(\mathbf{p}^i)))\nabla_{\theta}(\mathbf{t}^i)^{\top}(\mathbf{p}^i)\\
        & = -\sigma((\mathbf{t}^i)^{\top}(\mathbf{p}^i))\nabla_{\theta}((\mathbf{t}^i)^{\top}(\mathbf{p}^i))\\
        & = -\mathbf{J}_{\mathbf{x}^i}(\theta)\sigma((\mathbf{t}^i)^{\top}(\mathbf{p}^i))(\mathbf{p}^i_c (1-\mathbf{p}^i_c)  \mathbf{t}^i_c)\\
    \end{split}
\end{equation}

\section{Appendix: Dataset Description}

\begin{table}[H]
\centering
\resizebox{0.8\textwidth}{!}{%
\begin{tabular}{l|c|cc|c}
\toprule \hline
Training Scheme & NIH/CXP & OPI  & PDC & Number of Classes \\ \hline
NIH-OPI-PDC     & 83,672            & 2,971          & 14,714        & 14                \\
CXP-OPI-PDC    & 170,958           & 2,823         & 12,885        & 8                 \\ \hline \bottomrule
\end{tabular}%
}
\caption{Statistics of training/testing sets after trimming for consistency between different datasets.}
\label{tab:my-table}
\end{table}
In order to keep the classes consistent between different datasets, we align training set and testing set with the shared classes and exclude unique classes (see Tab.~\ref{tab:my-table}). For further detail, please check~\cite{Cohen2020xrv} for each class.

\section{Appendix: Synthetic Label Noise Result }
We include results using the public noisy-label medical image benchmark from~\cite{zhang2021alleviating} to test our method for different rates of symmetric noise. We tested our method on their ResNet-18 benchmark, where their baseline accuracy result using the 100\% clean set is 64.4\%. With 20\% symmetric noise, our method without our prior reaches 61.3\% and with our prior has 63.1\% (best result in~\cite{zhang2021alleviating}: 59.37\%). With 40\% symmetric noise, our method without our prior reaches 50.7\% and with our prior has 53.4\% (best result in~\cite{zhang2021alleviating}: 49.65\%). 
\section{Appendix: Memory Footprint}
We analysis the memory footprint of our method. In particular, our memory module is a matrix with N x C dimension. We used debugging tools to analyse our memory module and noted that we only required 4 MB of GPU memory, and at each iteration, we only backpropagated through a small subset of the memory.

\end{document}